# Strong Coresets for Hard and Soft Bregman Clustering with Applications to Exponential Family Mixtures

**Mario Lucic†**
ETH Zurich

**Olivier Bachem†**
ETH Zurich

**Andreas Krause**
ETH Zurich

## Abstract

*Coresets* are efficient representations of data sets such that models trained on the coreset are *provably* competitive with models trained on the original data set. As such, they have been successfully used to scale up clustering models such as K-Means and Gaussian mixture models to massive data sets. However, until now, the algorithms and the corresponding theory were usually specific to each clustering problem.

We propose a single, practical algorithm to construct strong coresets for a large class of hard and soft clustering problems based on Bregman divergences. This class includes hard clustering with popular distortion measures such as the Squared Euclidean distance, the Mahalanobis distance, KL-divergence and Itakura-Saito distance. The corresponding soft clustering problems are directly related to popular mixture models due to a dual relationship between Bregman divergences and Exponential family distributions. Our theoretical results further imply a randomized polynomial-time approximation scheme for hard clustering. We demonstrate the practicality of the proposed algorithm in an empirical evaluation.

## 1 INTRODUCTION

Clustering is the task of partitioning data points into groups (or clusters) such that similar data points are assigned to the same cluster. It is widely used in machine learning, data mining and information theory and remains an important unsupervised learning problem: According to Wu et al. (2008), *Lloyd's algorithm*, a local search algorithm for K-Means, is still one of the ten most popular algorithms for data mining.

Over the last several years, the world has witnessed the emergence of data sets of an unprecedented scale across different scientific disciplines. The large size of such data sets presents new computational challenges as existing algorithms can be computationally infeasible in the context of millions or even billions of data points. Clustering is not exempt from this problem: Popular algorithms to solve both hard clustering problems (such as K-Means) and soft clustering problems (such as *Gaussian mixture models*) typically require multiple passes through a data set.

A well established technique for scaling several clustering problems (including K-Means and GMMs) is based on *coresets* – a data summarization technique originating from computational geometry. A *coreset* is an efficient representation of the full data set such that models trained on a coreset are *provably* competitive with models trained on the original data set. Since coresets are typically small and easy to construct, they allow for efficient approximate inference with strong theoretical guarantees. However, until now, the algorithms and the corresponding theory were specific to each clustering problem.

*Bregman divergences* are a class of dissimilarity measures that are characterized by the property that the mean of a set of points is the optimal representative of that set. As such, the *Bregman hard clustering problem* generalizes a variety of important clustering problems such as K-Means and various information theoretic extensions (Banerjee et al., 2005). A family of corresponding soft clustering problems are denoted by *Bregman soft clustering* (Banerjee et al., 2005) and are closely related to fitting mixture models with exponential family distributions. Hence, *Bregman clustering* offers a natural framework for studying a variety of hard and soft clustering problems.

---

†These authors contributed equally to this work.

| DOMAIN | $\phi(t)$ | $\mu$ | $A$ | $d_\phi(p,q)$ | DIVERGENCE |
|---|---|---|---|---|---|
| $\mathbb{R}^d$ | $||t||_2^2$ | 1 | $I_d$ | $||p-q||_2^2$ | Squared Euclidean |
| $\mathbb{R}^d$ | $t^T A t$ | 1 | $A$ | $(p-q)^T A(p-q)$ | Mahalanobis |
| $[\lambda, \nu]^d \subseteq \mathbb{R}_+^d$ | $\sum t_i \ln(t_i) - t_i$ | $\frac{\lambda}{\nu}$ | $\frac{1}{2\lambda} I_d$ | $\sum p_i \ln p_i/q_i - \sum(p_i - q_i)$ | Relative Entropy |
| $[\lambda, \nu]^d \subseteq \mathbb{R}_+^d$ | $-\sum \ln(t_i)$ | $\frac{\nu^2}{\lambda^2}$ | $\frac{1}{2\lambda^2} I_d$ | $\sum p_i/q_i - \ln(p_i/q_i) - 1$ | Itakura-Saito |
| $[\lambda, \nu]^d \subseteq \mathbb{R}_+^d$ | $\sum 1/t_i^\alpha$ | $\frac{\alpha+2}{\nu^{\alpha+2}}$ | $\frac{\alpha(1-\alpha)}{2\lambda^{\alpha+2}} I_d$ | $\sum 1/p_i^\alpha - (\alpha + 1)/q_i^\alpha + \alpha p_i/q_i^{\alpha+1}$ | Harmonic ($\alpha > 0$) |
| $[\lambda, \nu]^d \subseteq \mathbb{R}_+^d$ | $\sum t_i^\alpha$ | $\frac{\lambda^{\alpha-2}}{\nu^{\alpha-2}}$ | $\frac{\alpha(1-\alpha)}{2}\nu^{\alpha-2} I_d$ | $\sum p_i^\alpha + (\alpha - 1)q_i^\alpha - \alpha p_i q_i^{\alpha-1}$ | Norm-like ($\alpha > 2$) |
| $[\lambda, \nu]^d \subseteq \mathbb{R}_+^d$ | $\sum e_i^t$ | $e^{-(\nu-\lambda)}$ | $\frac{e^\nu}{2} I_d$ | $\sum e^{p_i} - (p_i - q_i + 1)e^{q_i}$ | Exponential Loss |
| $[-\nu, \nu]^d \subseteq (-1,1)^d$ | $-\sum \sqrt{1-t_i^2}$ | $2(1-\nu^2)^{3/2}$ | $\frac{1}{2(1-\nu^2)^{3/2}} I_d$ | $\sum(1 - p_i q_i)/(\sqrt{1-q_i^2}) - \sqrt{1-p_i^2}$ | Hellinger |

Table 1: Selection of $\mu$-similar Bregman divergences (Ackermann and Blömer, 2009).

**Our contributions.** In this paper, we consider the problem of scaling Bregman clustering to massive data sets using coresets. As our key contributions, we:

- provide strong coresets of size *independent of the data set size* for *all* $\mu$-similar Bregman divergences,

- prove that the *same* practical coreset construction works for both hard and soft clustering problems,

- provide a randomized polynomial-time approximation scheme for the corresponding hard clustering problems,

- establish the combinatorial complexity of mixture models induced by the regular exponential family,

- demonstrate the practicality of the proposed algorithm in an empirical evaluation.

## 2 BACKGROUND

**Bregman divergence.** For any strictly convex, differentiable function $\phi : \mathcal{K} \to \mathbb{R}$, the Bregman divergence with respect to $\phi$ for all $p, q \in \mathcal{K}$ is defined as

$$d_\phi(p,q) = \phi(p) - \phi(q) - \nabla\phi(q)^T(p-q).$$

For example, setting $\mathcal{K} = \mathbb{R}^d$ and $\phi(q) = ||q||_2^2$ results in $d_\phi(p,q) = ||p-q||_2^2$ which is the squared Euclidean distance. Bregman divergences are characterized by the fact that the mean of a set of points minimizes the sum of Bregman divergences between these points and any other point. More formally, for any Bregman divergence $d_\phi$ and any finite set $\mathcal{X} \subseteq \mathcal{K}$, it holds that

$$\arg\min_{z\in\mathbb{R}^d} \sum_{x\in\mathcal{X}} d_\phi(x,z) = \frac{1}{|\mathcal{X}|}\sum_{x\in\mathcal{X}} x. \qquad (1)$$

Furthermore, Banerjee et al. (2005) proved that any function satisfying (1) is a Bregman divergence.

$\mu$-similar Bregman divergences are a subclass of Bregman divergences related to the squared Mahalanobis distance.[1] The class of $\mu$-similar Bregman divergences includes many popular Bregman divergences such as the squared Euclidean distance, the squared Mahalanobis distance, the KL-divergence and the Itakura-Saito distance. Several other divergences from this class are shown in Table 1.

**Definition 1 ($\mu$-similar Bregman divergence)**
*A Bregman divergence* $d_\phi$ *on domain* $\mathcal{K} \subseteq \mathbb{R}^d$ *is* $\mu$-*similar for some* $\mu > 0$ *iff there exists a positive definite matrix* $A \in \mathbb{R}^{d\times d}$ *such that, for each* $p, q \in \mathcal{K}$,

$$\mu\, d_A(p,q) \leq d_\phi(p,q) \leq d_A(p,q)$$

*where* $d_A$ *denotes the squared Mahalanobis distance.*

**Coresets.** A coreset is a weighted subset of the data such that the quality of any clustering evaluated on the coreset closely approximates the quality on the full data set. Consider a cost function that depends on a set of points $\mathcal{X}$ and a query $Q \in \mathcal{Q}$ that is additively decomposable into non-negative functions $\{f_Q(x)\}_{x\in\mathcal{X}}$, i.e.

$$\text{cost}(\mathcal{X}, Q) = \sum_{x\in\mathcal{X}} f_Q(x).$$

For example, for the K-Means clustering problem, $f_Q(x)$ is the squared Euclidean distance of $x$ to the closest cluster center in the set $Q$. The key idea behind coresets is to find a weighted subset $\mathcal{C}$ such that the cost of a query $Q$ can be approximated on $\mathcal{C}$ by

$$\text{cost}(\mathcal{C}, Q) = \sum_{(w,c)\in\mathcal{C}} w f_Q(c).$$

A weighted subset $\mathcal{C}$ is an $\varepsilon$-coreset of $\mathcal{X}$ if it approximates the cost function of the full data set up to a multiplicative factor of $1 \pm \varepsilon$ uniformly for *all* queries $Q \in \mathcal{Q}$, i.e.

$$(1-\varepsilon)\,\text{cost}(\mathcal{X}, Q) \leq \text{cost}(\mathcal{C}, Q) \leq (1+\varepsilon)\,\text{cost}(\mathcal{X}, Q).$$

Since the cost contributions $f_Q(x)$ and the space of queries $\mathcal{Q}$ depend on the problem at hand, coresets are inherently problem-specific.

---

[1] The squared Mahalanobis distance $d_A$ for any positive definite matrix $A \in \mathbb{R}^{d\times d}$ and any $p, q \in \mathbb{R}^d$ is defined as $d_A(p,q) = (p-q)^T A(p-q)$.

Coresets have been the subject of several recent papers, with focus on unsupervised parametric and nonparametric models (Feldman et al., 2011; Balcan et al., 2013; Bachem et al., 2015; Lucic et al., 2015), as well as in the context of empirical risk minimization (Reddi et al., 2015).

As the coreset property bounds the approximation error for all queries, the difference between the solution on the full data set and the solution on the coreset is bounded. Hence, one can use any solver on the coreset instead of the full data set and obtain provable approximation guarantees.[2] At the same time, coresets are usually sublinear in (or even independent of) the number of samples which implies that one can run computationally intensive algorithms that would otherwise be infeasible.

Additionally, coresets are a practical and flexible tool that requires no assumptions on the data. While the theory behind coresets requires elaborate tools from computational geometry, the resulting coreset construction algorithms are simple to implement.

A key property of coresets is that they can be constructed both in a distributed and a streaming setting. The constructions rely on the property that both unions of coresets and coresets of coresets are coresets (Har-Peled and Mazumdar, 2004). In fact, Feldman et al. (2013) use these properties to construct coresets in a tree-wise fashion which can be parallelized in a Map-Reduce style or used to maintain an up-to-date coreset in a streaming setting by applying the *static-to-dynamic transformation* (Bentley and Saxe, 1980).

## 3 STRONG CORESETS FOR HARD CLUSTERING

The goal of the K-Means clustering problem is to find $k$ cluster centers $C$ such that the quantization error

$$\text{cost}_{\texttt{kmeans}}(\mathcal{X}, C) = \sum_{x \in \mathcal{X}} \min_{c \in C} ||x - c||_2^2$$

is minimized, where $\mathcal{X} \subseteq \mathbb{R}^d$ denotes the set of data points to be clustered. By replacing the squared Euclidean distance $|| \cdot ||_2^2$ with a Bregman divergence $\text{d}_\phi$, K-Means clustering generalizes to *Bregman hard clustering*. In this problem the goal is to compute a set of $k$ cluster centers $C$ minimizing

$$\text{cost}_{\text{H}}(\mathcal{X}, C) = \frac{1}{|\mathcal{X}|} \sum_{x \in \mathcal{X}} \text{d}_\phi(x, C) \qquad (2)$$

---

---

**Algorithm 1** Bregman hard $k$-clustering with $\text{d}_\phi$

**Require:** $\mathcal{X}, k$, initial centers $\{\mu_j\}_{j=1}^k$
1: **repeat**
2:   **for** $j \leftarrow 1$ **to** $k$
3:     $\mathcal{X}_j \leftarrow \emptyset$
4:   **for** $i \leftarrow 1$ **to** $n$
5:     $j \leftarrow \arg\min_{1 \leq \ell \leq k} \text{d}_\phi(x_i, \mu_\ell)$
6:     $\mathcal{X}_j \leftarrow \mathcal{X}_j \cup \{x_i\}$
7:   **for** $j \leftarrow 1$ **to** $k$
8:     $\mu_j \leftarrow \frac{1}{|\mathcal{X}_j|} \sum_{x \in \mathcal{X}_j} x$
9: **until** convergence
10: **return** $\{\mu_j\}_{j=1}^k$

---

where we define $\text{d}_\phi(x, C) = \min_{c \in C} \text{d}_\phi(x, c)$. For weighted data sets, the contribution of a point to the cost function is scaled by its weight $\nu(x)$.

The notion of hard clustering relates to the fact that, as in K-Means clustering, the *minimum* with regards to the set of cluster centers $C$ leads to a hard assignment of each data point to its closest cluster center. Furthermore, the assignment boundary between two arbitrary centers is a $(d-1)$-dimensional hyperplane and the set of cluster centers $C$ induces a Voronoi partitioning on the data set $\mathcal{X}$ (Banerjee et al., 2005).

Lloyd's algorithm (Lloyd, 1982) solves the K-Means problem by iterating between assigning points to closest cluster centers and recalculating the centers as the mean of the assigned points. As the mean is an optimal representative of a single cluster for any Bregman divergence, Lloyd's algorithm can be naturally generalized to the *Bregman hard clustering algorithm* detailed in Algorithm 1 (Banerjee et al., 2005).

Each iteration of Algorithm 1 incurs a computational cost of $\mathcal{O}(nkd)$. As the number of iterations until convergence can be very large, coresets can be used to scale Bregman hard clustering. Previous approaches either only considered *weak* coresets (Ackermann and Blömer, 2009) or impose prior assumptions on the data (Feldman et al., 2013). In contrast, we provide a significantly stronger theoretical guarantee via *strong* coresets, i.e. coresets for which the approximation guarantee holds for *all* possible queries.

**Definition 2 (Coreset definition)** *Let $\varepsilon > 0$ and $k \in \mathbb{N}$. Let $\text{d}_\phi$ be a $\mu$-similar Bregman divergence on domain $\mathcal{K}$ and $\mathcal{X} \subseteq \mathcal{K}$ be a finite set of points. The weighted set $\mathcal{C}$ is an $(\varepsilon, k)$ coreset of $\mathcal{X}$ for hard clustering with $\text{d}_\phi$ if for any set $Q \subseteq \mathcal{K}$ of cardinality $k$*

$$|\text{cost}_{\text{H}}(\mathcal{X}, Q) - \text{cost}_{\text{H}}(\mathcal{C}, Q)| \leq \varepsilon \, \text{cost}_{\text{H}}(\mathcal{X}, Q).$$

### 3.1 Coreset construction algorithm

The idea behind the proposed coreset construction is straightforward: The objective function for Bregman hard clustering in (2) is additively decomposable over $\mathcal{X}$ and the contribution of each point is independent of all the other points. Hence, the cost evaluated on a uniform subsample of the data points is an unbiased estimator of the true objective function.

While this intuition is simple, unbiasedness is not sufficient and uniform subsampling does not provide the strong theoretical guarantee required by Definition 2: Single points can potentially have a large impact on the objective function and force the sample size to $\Omega(n)$.[3] In particular, this occurs if the clusters are imbalanced or if there are points far away from the bulk of the data. Moreover, the coreset property in Definition 2 needs to hold uniformly for all queries in the (possibly infinite) set $Q$.

To obtain the strong coreset property, we propose a coreset construction based on the *importance sampling* framework by Langberg and Schulman (2010) and Feldman and Langberg (2011). Consider a Bregman hard clustering problem defined by a $\mu$-similar Bregman divergence $d_\phi$ on domain $\mathcal{K}$ and a finite data set $\mathcal{X} \subseteq \mathcal{K}$. The construction consists of two steps:

**Step 1.** We first find a rough approximation (*bicriteria approximation*) of the optimal clustering. As $\mu$-similar Bregman divergences are closely related to the squared Mahalanobis distance, we show in the proof of Theorem 1 that it is sufficient to find a rough approximation with regards to the Mahalanobis distance. To this end, we apply $D^2$-sampling which was previously analyzed by Arthur and Vassilvitskii (2007) in the context of the popular `k-means++` algorithm. The idea is to sample data points as cluster centers using an adaptive sampling scheme: the first cluster center is sampled uniformly at random and additional points are then iteratively sampled with probability proportional to the minimum squared Mahalanobis distance to the already sampled cluster centers. We prove that this procedure detailed in Algorithm 2 produces a solution that is, with constant probability, $\mathcal{O}(\log k)$ competitive with the optimal clustering in terms of Mahalanobis distance. Under natural assumptions on the data, a bicriteria approximation can even be computed in sublinear time (Bachem et al., 2016).

**Step 2.** The rough approximation is then used in Algorithm 3 to compute an importance sampling distribution. The idea is to sample points with a potentially

---

**Algorithm 2** Mahalanobis $D^2$-sampling

**Require:** $\mathcal{X}$, $k$, $d_A$
1: Uniformly sample $x \in \mathcal{X}$ and set $B = \{x\}$.
2: **for** $i \leftarrow 2, 3, \ldots, k$
3:     Sample $x \in \mathcal{X}$ with probability $\frac{d_A(x,B)}{\sum_{x' \in \mathcal{X}} d_A(x',B)}$ and add it to $B$.
4: **return** $B$

---

**Algorithm 3** Coreset construction

**Require:** $\mathcal{X}$, $k$, $B$, $m$, $d_A$
1: $\alpha \leftarrow 16(\log k + 2)$
2: **for each** $b_i$ in $B$
3:     $B_i \leftarrow$ Set of points from $\mathcal{X}$ closest to $b_i$ in terms of $d_A$. Ties broken arbitrarily.
4: $c_\phi \leftarrow \frac{1}{|\mathcal{X}|} \sum_{x' \in \mathcal{X}} d_A(x', B)$
5: **for each** $b_i \in B$ and $x \in B_i$
6:     $s(x) \leftarrow \frac{\alpha\, d_A(x,B)}{c_\phi} + \frac{2\alpha \sum_{x' \in B_i} d_A(x',B)}{|B_i| c_\phi} + \frac{4|\mathcal{X}|}{|B_i|}$
7: **for each** $x \in \mathcal{X}$
8:     $p(x) \leftarrow s(x) / \sum_{x' \in \mathcal{X}} s(x')$
9: $\mathcal{C} \leftarrow$ Sample $m$ weighted points from $\mathcal{X}$ where each point $x$ has weight $\frac{1}{mp(x)}$ and is sampled with probability $p(x)$.
10: **return** $\mathcal{C}$

---

high impact on the objective more frequently but assign them a lower weight. The sensitivity $s(x)$ of a point $x \in \mathcal{X}$ is the maximal ratio between the cost contribution of that point and the average contribution of all points (Langberg and Schulman, 2010), i.e.

$$\sigma(x) = \max_{Q \subseteq \mathcal{K}: |Q| = k} \frac{d_\phi(x, Q)}{\frac{1}{|\mathcal{X}|} \sum_{x' \in \mathcal{X}} d_\phi(x', Q)}.$$

We derive an upper bound for $\sigma(x)$ and use it as the sampling distribution in Algorithm 3. Intuitively, this specific choice bounds the variance of the importance sampling scheme (Feldman and Langberg, 2011) and, if we sample enough points, produces a coreset. This result is formally stated and proven in Theorem 1 where we provide a bound on the required coreset size.

Finally, we can solve the Bregman hard clustering problem on the coreset using a weighted version of the Bregman hard clustering algorithm.

### 3.2 Analysis

Our main result is that this construction leads to valid coresets for the Bregman hard clustering problem. In particular, the required coreset size does not depend on the size $n$ of the original data set.

---

**Theorem 1** *Let $\varepsilon \in (0, 1/4)$, $\delta > 0$ and $k \in \mathbb{N}$. Let $d_\phi$ be a $\mu$-similar Bregman divergence on domain $\mathcal{K}$ and denote by $d_A$ the corresponding squared Mahalanobis* distance. Let $\mathcal{X}$ be a set of points in $\mathcal{K}$ and let $B \subseteq \mathcal{X}$ be the set with the smallest quantization error in terms of $\mathrm{d}_A$ among $\mathcal{O}(\log \frac{1}{\delta})$ runs of Algorithm 2. Let $C$ be the output of Algorithm 3 with

$$m = \mathcal{O}\left(\frac{dk^3 + k^2 \log \frac{1}{\delta}}{\mu^2 \varepsilon^2}\right).$$

---

[3]A simple example is a data set where a first cluster contains $n - 1$ points at a single location and a second cluster consists of one point arbitrarily far away from the first cluster.

Then, with probability at least $1 - \delta$, the set $\mathcal{C}$ is a $(\varepsilon, k)$-coreset of $\mathcal{X}$ for hard clustering with $\mathrm{d}_\phi$.

The proof is provided in Section C of the supplementary material. The main steps include bounding the sensitivity in Lemma 2 and bounding the combinatorial complexity of Bregman hard clustering in Theorem 6. In practice, it is usually sufficient to run Algorithm 2 only once and to fix the coreset size $m$ instead of $\varepsilon$ (see Section 5).

As an immediate consequence of the coreset property, the optimal clustering obtained on the coreset is provably competitive with the optimal clustering on the full data set when evaluated on the full data set.

**Lemma 1** *Let $\varepsilon \in (0, 1)$ and let $\mathrm{d}_\phi$ be a $\mu$-similar Bregman divergence on domain $\mathcal{K}$. Let $\mathcal{X} \subseteq \mathcal{K}$ be a data set, $k \in \mathbb{N}$ and $\mathcal{C}$ be an $(\varepsilon/3, k)$-coreset of $\mathcal{X}$ for hard clustering with $\mathrm{d}_\phi$. Let $Q^*_{\mathcal{X}}$ and $Q^*_{\mathcal{C}}$ denote the optimal set of cluster centers for $\mathcal{X}$ and $\mathcal{C}$ respectively. Then,*

$$\mathrm{cost}_\mathrm{H}(\mathcal{X}, Q^*_{\mathcal{C}}) \leq (1 + \varepsilon) \, \mathrm{cost}_\mathrm{H}(\mathcal{X}, Q^*_{\mathcal{X}}).$$

The proof is presented in the Section C of the supplementary material.

### 3.3 Randomized polynomial-time approximation scheme

The fact that the size of the proposed coresets is independent of the number of data points $n$ readily implies a randomized polynomial-time approximation scheme (PTAS) for Bregman hard clustering with $\mu$-similar Bregman divergences. We first generate a coreset using Algorithm 3 and then consider all possible $k$ partitionings of the coreset points. By the coreset property, it is guaranteed that the centers of the best partitioning are $1 + \varepsilon$ competitive with the optimal solution.

**Theorem 2** *Let $\varepsilon \in (0, 3/4)$, $\delta > 0$ and let $\mathrm{d}_\phi$ be a $\mu$-similar Bregman divergence on domain $\mathcal{K}$. Let $\mathcal{X} \subseteq \mathcal{K}$ be a data set, $k \in \mathbb{N}$ and $\varepsilon$ the desired approximation error. Let $Q^*$ be the best solution from $\mathcal{O}(\log \frac{1}{\delta})$ runs of Algorithm 4. Then, with probability at least $1 - \delta$,*

$$\mathrm{cost}_\mathrm{H}(\mathcal{X}, Q^*) \leq (1 + \varepsilon) \min_Q \mathrm{cost}_\mathrm{H}(\mathcal{X}, Q).$$

*The time complexity is $\mathcal{O}\left((nkd + 2^{\mathrm{poly}(kd/\mu\varepsilon)}) \log \frac{1}{\delta}\right)$.*

---

**Algorithm 4** Randomized PTAS

**Require:** $\mathcal{X}$, $k$, $\varepsilon$, $\mathrm{d}_\phi$
1: $\mathcal{C} \leftarrow (k, \varepsilon/3)$-coreset for $\mathcal{X}$ with respect to $\mathrm{d}_\phi$.
2: $\mathcal{P} \leftarrow$ Centers of all $k$-partitionings of $\mathcal{C}$.
3: $Q^* \leftarrow \arg\min_{P \in \mathcal{P}} \frac{1}{|\mathcal{C}|} \sum_{(w,c) \in \mathcal{C}} w \, \mathrm{d}_\phi(c, P)$
4: **return** $Q^*$

---

The correctness follows from Lemma 1 and the fact that that the number of $k$ partitionings of the coreset points is independent of $n$. As this algorithm is primarily of theoretical interest, we recommend running the approach presented in Section 3.1 to solve Bregman clustering problems in practice.

## 4 STRONG CORESETS FOR SOFT CLUSTERING

In hard clustering each data point is assigned to exactly one cluster. In contrast, in soft clustering each data point is assigned to each cluster with a certain probability. A prototypical example of soft clustering is fitting the parameters of a Gaussian mixture model in which one assumes that all data points are generated from a mixture of a finite number of Gaussians with unknown parameters. Other popular models include the Poisson mixture model, the mixture of multinomials and the mixture of exponentials.

Banerjee et al. (2005) show that there is a bijection between regular exponential family distributions and Bregman divergences. In particular, the log-likelihood of exponential family mixture models can be expressed in terms of Bregman divergences (see Section 4.2). By considering the resulting objective, one obtains *Bregman soft clustering*. The intuition is that Bregman hard clustering can be turned into a soft clustering problem by replacing the *min* function by a *soft-min* function. More formally, let $\mathrm{d}_\phi$ be a Bregman divergence and $\mathcal{X} \subseteq \mathcal{K}$ be a set of $n$ points. Let $k \in \mathbb{N}$, $w = (w_1, \dots, w_k) \subseteq \mathbb{R}^k$ and $\theta = (\theta_1, \dots, \theta_k) \subset \mathbb{R}^{kd}$ and let $Q$ be the concatenation of $w$ and $\theta$. The goal of Bregman soft clustering is to minimize

$$\mathrm{cost}_\mathrm{s}(\mathcal{X}, Q) = -\sum_{i=1}^{n} \ln\left(\sum_{j=1}^{k} w_j \exp(-d_\phi(x_i, \theta_j))\right)$$

with respect to $Q$ under the constraint that $w_j > 0$, $1 \leq j \leq k$ and $\sum_{j=1}^{k} w_j = 1$. Similar to Bregman hard clustering, the soft clustering problem can be solved using an *expectation-maximization* algorithm (Banerjee et al., 2005) which is detailed in Algorithm 5. The main difference with respect to hard clustering is that a probability distribution over assignments of points to clusters is maintained.

**Algorithm 5** Bregman soft clustering with $d_\phi$

**Require:** $\mathcal{X}, k$, initial parameters $\{(\theta_j, w_j)\}_{j=1}^k$
1: **repeat**
2:     **for** $i = 1$ to $n$
3:         **for** $j = 1$ to $k$
4:             $\eta_{ij} = \frac{w_j \exp(-d_\phi(x_i, \theta_j))}{\sum_{\ell=1}^k w_\ell \exp(-d_\phi(x_i, \theta_\ell))}$
5:     **for** $j = 1$ to $k$
6:         $w_j \leftarrow \frac{1}{n} \sum_{i=1}^n \eta_{ij}$
7:         $\theta_j \leftarrow \frac{\sum_{i=1}^n \eta_{ij} x_i}{\sum_{i=1}^n \eta_{ij}}$
8: **until** convergence
9: **return** $\{\theta_j, w_j\}_{j=1}^k$

### 4.1 Coresets for Bregman soft clustering

As in the hard clustering case, we define the coreset property in terms of a cost function with respect to some set of queries. In the case of soft clustering, the queries are the parameters of the mixture model.

**Definition 3** *Let $\varepsilon > 0$ and $k \in \mathbb{N}$. Let $d_\phi$ be a $\mu$-similar Bregman divergence on domain $\mathcal{K}$ and $\mathcal{X} \subseteq \mathcal{K}$ be a set of points. The weighted set $\mathcal{C}$ is an $(\varepsilon, k)$ coreset of $\mathcal{X}$ for soft clustering with $d_\phi$ if for any set $Q \subseteq \mathcal{K}$ of cardinality $k$*

$$|\text{cost}_s(\mathcal{X}, Q) - \text{cost}_s(\mathcal{C}, Q)| \leq \varepsilon \, \text{cost}_s(\mathcal{X}, Q).$$

We prove that the same coreset construction provided in Section 3.1 for hard clustering also computes valid coresets for the soft clustering problem. However, due to the higher combinatorial complexity of the underlying function space, more points need to be sampled to guarantee that the resulting weighted set is a coreset.

**Theorem 3** *Let $\varepsilon \in (0, 1/4), \delta > 0$ and $k \in \mathbb{N}$. Let $d_\phi$ be a $\mu$-similar Bregman divergence on domain $\mathcal{K}$ and denote by $d_A$ the corresponding squared Mahalanobis distance. Let $\mathcal{X}$ be a set of points in $\mathcal{K}$ and let $B \subseteq \mathcal{X}$ be the set with the smallest quantization error in terms of $d_A$ among $\mathcal{O}(\log \frac{1}{\delta})$ runs of Algorithm 2. Let $\mathcal{C}$ be the output of Algorithm 3 with coreset size*

$$m = \mathcal{O}\left(\frac{d^2 k^4 + k^2 \log \frac{1}{\delta}}{\mu^2 \varepsilon^2}\right).$$

*Then, with probability at least $1 - \delta$, the set $\mathcal{C}$ is a $(\varepsilon, k)$-coreset of $\mathcal{X}$ for soft clustering with $d_\phi$.*

The proof which includes a bound on the combinatorial complexity of mixtures of regular exponential family distributions is provided in Section D of the supplementary material.

### 4.2 Estimation for exponential family mixtures

Finding the maximum likelihood estimate of a set of parameters for a single exponential family can be done efficiently trough the use of sufficient statistics. However, fitting the parameters of a mixture is a notoriously hard task – the mixture is not in the exponential family. More formally, consider a set $\tilde{\mathcal{X}}$ of $n$ points drawn independently from a stochastic source that is a mixture of $k$ densities of the same exponential family. Given $\tilde{\mathcal{X}}$, we would like to estimate the parameters $\{w_j, \theta_j\}_{j=1}^k$ of the mixture model using maximum likelihood estimation, i.e.

$$\max_{\{w_j, \theta_j\}_{j=1}^k} \mathcal{L}(\tilde{\mathcal{X}} \mid \theta) = \sum_{i=1}^n \ln\left(\sum_{j=1}^k w_j p_\psi(\tilde{x}_i \mid \theta_j)\right). \quad (3)$$

As shown by Banerjee et al. (2005) there is a bijection between regular exponential families and regular Bregman divergences that allows us to rewrite (3) as

$$\mathcal{L}(\mathcal{X} \mid Q) = \sum_{i=1}^n \ln\left(\sum_{j=1}^k w_j \exp(-d_\phi(x, \eta_j)) b_\phi(x)\right)$$
$$= \sum_{i=1}^n \ln(b_\phi(x_i))$$
$$+ \sum_{i=1}^n \ln\left(\sum_{j=1}^k w_j \exp(-d_\phi(x_i, \eta_j))\right)$$

where $d_\phi$ is the corresponding Bregman divergence and both $\tilde{x}$ and $\eta_j$ are related to $x$ and $\theta_j$ by Legendre duality via $\phi$. Since the first summand is independent of the model parameters, maximizing the likelihood of the mixture model is equivalent to minimizing

$$\text{cost}_s(\mathcal{X}, Q) = -\sum_{i=1}^n \ln\left(\sum_{j=1}^k w_j \exp(-d_\phi(x_i, \theta_j))\right).$$

This is precisely the cost function of the coreset defined in Section 4.1. Hence, our coreset construction can be used for maximum likelihood estimation of regular exponential family mixtures by first constructing a coreset and then applying Bregman soft clustering (Algorithm 5) on the coreset. Moreover, as $\varepsilon \to 0$

$$|\mathcal{L}(\mathcal{X}|Q) - \mathcal{L}(\mathcal{C}|Q)| = |\text{cost}_s(\mathcal{X}, Q) - \text{cost}_s(\mathcal{C}, Q)| \to 0$$

uniformly over $Q \in \mathcal{Q}$. As a byproduct of our theoretical analysis, we also provide a bound of $\mathcal{O}(k^4 d^2)$ on the combinatorial complexity of mixtures of regular exponential family distributions with $k$ components and $d$ dimensions (see Theorem 7 in Section D of the supplementary material).

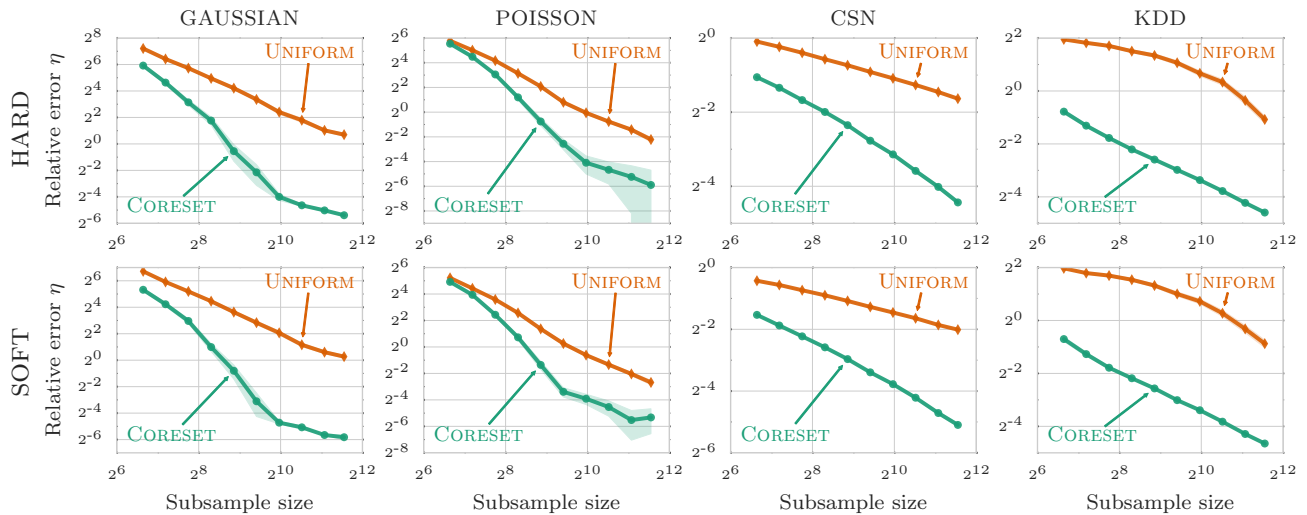

Figure 1: Relative error in relation to subsample size. For a fixed subsample size, coresets strongly outperform uniform subsampling. Shaded areas denote confidence intervals based on 500 independent trials.

## 5  EXPERIMENTAL EVALUATION

In this section, we demonstrate that our proposed core-set construction can be used in practice to speed up both hard and soft clustering clustering problems. We compare our proposed coreset construction[4] to solving the clustering problem on both uniform subsamples of the data set and the original data set itself.

We first generate a weighted subsample of the data using either uniform subsampling or the proposed coreset construction. To solve the corresponding hard (soft) clustering problem on the subsample we apply Algorithm 1 (5) with adaptive seeding using $D^2$-sampling. We then evaluate the computed solution on the full data set to obtain the cost $C_{ss}$ and measure the CPU time elapsed for both the subsampling and the solving step. We average the results over $r = 500$ independent trials. Independently, we also measure the CPU time elapsed and the solution quality $C_{full}$ obtained when training using the full data set, again averaged across $r = 500$ independent trials. Finally, we calculate the *relative error* $\eta = (C_{ss} - C_{full})/C_{full}$ for both uniform subsampling and coresets.

### 5.1  Data sets and parameters

Following Banerjee et al. (2005) we use (regular) exponential family mixture models to generate two synthetic data sets. We sample the model parameters from the associated conjugate prior and cluster the data using the corresponding dual Bregman divergence (as detailed in Section 4.2).

GAUSSIAN. This data set consists of 10,000 points which are drawn from a mixture of $k = 50$ isotropic Gaussians in $d = 10$ dimensions. We use a $k$-dimensional Dirichlet distribution with concentration parameter $\alpha = 0.5$ to sample the mixture weights. The means of each of the $k$ components are in turn sampled from a Gaussian with zero mean and variance of 5000. We solve both hard and soft clustering problems with the squared Euclidean distance as the Bregman divergence and $k = 50$ cluster centers.

POISSON. This data set consists of 10,000 points drawn from a mixture of $k = 50$ multivariate Poisson distributions in $d = 10$ dimensions. In a given component $j$, each dimension is independently sampled, i.e. $x_i \sim \text{Poi}(\mu_{i,j})$ for each $i = 1, 2, \ldots d$. For each component $j$ and dimension $i$, the parameter $\mu_{i,j}$ is sampled from a Gamma distribution with shape $\alpha = 10$ and rate $\beta = 10^{-3}$. We consider $k = 50$ cluster centers and use the relative entropy as the Bregman divergence.

CSN. In the Community Seismic Network (Faulkner et al., 2011) more than 7GB of cellphone accelerometer data were gathered and used to detect earthquakes. The data consists of 80,000 observations and 17 extracted features. We consider $k = 50$ cluster centers and use the squared Mahalanobis distance where $A$ is the inverse of the covariance matrix.

KDD. This data set was used in the Protein Homology Prediction KDD competition and contains 145,751 training examples with 74 features that measure the match between a protein and a native sequence. We consider $k = 50$ cluster centers and use the squared Euclidean distance as the Bregman Divergence.

---

[4] For the proposed coreset construction, we use $A = I$ for all data sets except for KDD where we use the inverse of the covariance matrix.

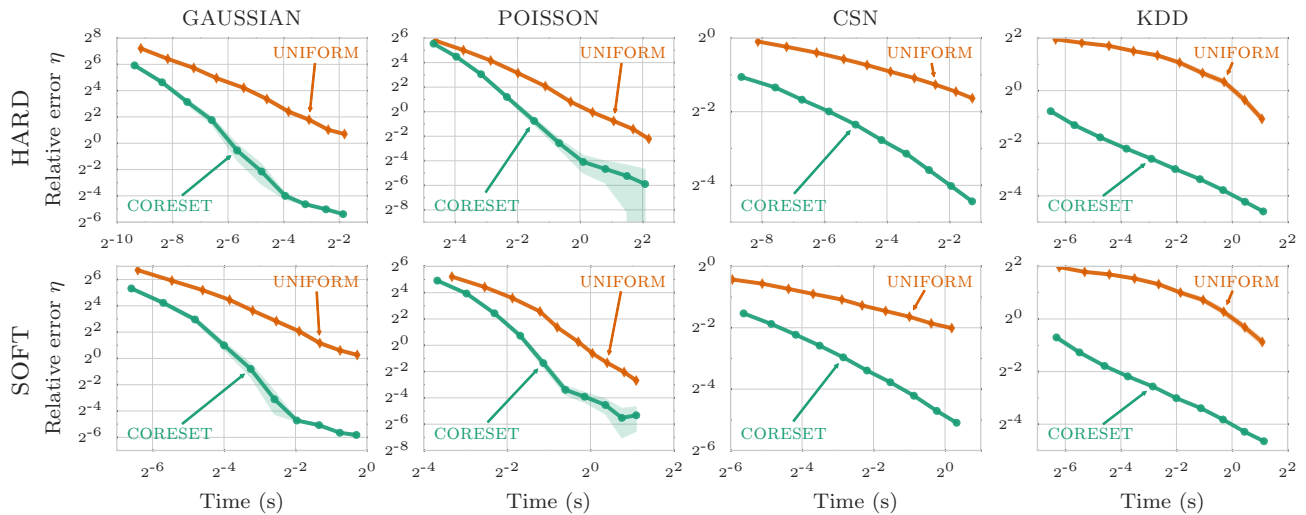

Figure 2: Time required to reach a fixed relative error using coresets and uniform subsampling. In all settings, coresets outperform uniform subsampling. Shaded areas denote confidence intervals based on 500 trials.

### 5.2 Discussion

Figure 1 shows the relative error $\eta$ for different subsample sizes. For all data sets and both hard and soft clustering, the relative error decreases as we increase the number of points in the subsample and even for relatively small subsample sizes our coreset construction produces competitive solution compared to solving on the full data set. Across all data sets, the proposed coreset construction outperforms uniform subsampling and achieves a lower relative cost for a given subsample size. Figure 2 shows that coresets reach a given relative error faster than uniform subsamples (even if we account for the time required to construct the coreset). The practical relevance can be seen in Table 2. For KDD and a subsample size of $s = 3000$, we obtain a speedup of $81.3\times$ using coresets, while only incurring a relative error of $4.1\%$. At the same time, uniform subsampling leads to a relative error of $47.4\%$.

Table 2: Hard clustering on KDD (3000 subsamples)

|  | Uniform | Coreset | Full |
|---|---|---|---|
| Time (s) | 2.11 | 2.17 | 176.84 |
| Speedup | 83.9x | 81.3x | 1.0x |
| Cost ($10^9$) | 290.46 | 205.18 | 197.02 |
| Relative error $\eta$ | 47.4% | 4.1% | 0.0% |

## 6 OTHER RELATED WORK

Relatively few approaches were suggested for scalable clustering with Bregman divergences. Ackermann and Blömer (2009) construct weak coresets for approximate clustering with $\mu$-similar Bregman divergences and obtain weak coresets of size $\mathcal{O}\left(\frac{1}{\varepsilon^2}k\log(n)\log(k|\Gamma|^k\log n)\right)$ where $|\Gamma| \leq n^{d\cdot\text{poly}(k/\varepsilon)}$.

For weak coresets, the approximation guarantee holds only for queries close to the optimal solution. As such, their applicability is severely limited. In contrast, we provide strong coresets for all $\mu$-similar Bregman divergences and generalize the results to the soft clustering case. Feldman et al. (2013) provide a coreset construction (albeit, the coreset is a set of clustering features, rather than a weighted subset of the original data set) for hard clustering with $\mu$-similar Bregman divergences with the additional restriction on the convex set $S$ (domain of $\phi$): every pair $p, q$ of points from $P$, $S$ must contain all points within a ball of radius $(4/m\varepsilon)d(p, q)$ around $p$ for a constant $\mu$. In contrast, our approach makes no such assumptions and provides a strong coreset with improved dependency on $\varepsilon$. Feldman et al. (2011) provide a coreset construction for the specific case of a mixture of semi-spherical Gaussians (bounded eigenvalues) and obtain a coreset of the size independent of the data set size.

## 7 CONCLUSION

We propose a single coreset construction algorithm for both hard and soft clustering with any $\mu$-similar Bregman divergence. As a separate result, we establish the combinatorial complexity of mixture models induced by the regular exponential family. Experimental results demonstrate the practical utility of the proposed approach. In particular, the coresets outperform uniform subsampling and enjoy speedups of several orders of magnitude compared to solving on the full data set while retaining guarantees on the approximation error.

**Acknowledgements**

This research was partially supported by ERC StG 307036 and the Zurich Information Security Center.

# A BACKGROUND

**Definition 4 (Sensitivity)** *Let $\mathcal{X}$ be a finite set of cardinality $n$ and denote by $f_x(Q)$ a cost function from $\mathcal{Q} \times \mathcal{X}$ to $[0, \infty)$. Define $\bar{f}_Q = \frac{1}{n} \sum_{x \in \mathcal{X}} f_x(Q)$ for all $Q \in \mathcal{Q}$. The sensitivity of the point $x \in \mathcal{X}$ with respect to a family of queries $\mathcal{Q}$ is defined as*

$$\sigma_{\mathcal{Q}}(x) = \max_{Q \in \mathcal{Q}} \frac{f_x(Q)}{\bar{f}_Q}.$$

**Definition 5 (Feldman and Langberg (2011))**
*Let $\mathcal{X}$ be a finite set of cardinality $n$ and denote by $f_x(Q)$ a cost function from $\mathcal{Q} \times \mathcal{X}$ to $[0, \infty)$. Define the set of functions $F = \{f_x(Q) \mid x \in \mathcal{X}\}$ from the set $\mathcal{Q}$ to $[0, \infty)$. The dimension $\dim(F)$ of $F$ is the minimum integer $d$ such that*

$$\forall S \subseteq F : |\{S \cap R \mid R \in \mathbf{ranges}(F)\}| \leq (|S| + 1)^d$$

*where $\mathbf{ranges}(F) = \{\mathbf{range}(Q, r) \mid Q \in \mathcal{Q}, r \geq 0\}$ and $\mathbf{range}(Q, r) = \{f \in F \mid f(Q) \leq r\}$ for every $Q \in \mathcal{Q}$ and $r \geq 0$.*

**Theorem 4 (Feldman and Langberg (2011))**
*Let $\epsilon \in (0, 1/4)$. Let $\mathcal{X}$ be a finite set of cardinality $n$ and denote by $f_x(Q)$ a cost function from $\mathcal{Q} \times \mathcal{X}$ to $[0, \infty)$. Define the set of functions $F = \{f_x(Q) \mid x \in \mathcal{X}\}$ from the set $\mathcal{Q}$ to $[0, \infty)$. Let $s : \mathcal{X} \to \mathbb{N} \setminus \{0\}$ be a function such that*

$$s(x) \geq \sigma_{\mathcal{Q}}(x), \quad \forall x \in \mathcal{X}$$

*and let $S = \sum_{x \in \mathcal{X}} s(x)/n$. For each $x \in \mathcal{X}$, let $g_x : \mathcal{Q} \to [0, \infty)$ be defined as $g_x(Q) = f_x(Q)/s(x)$. Let $G_x$ consist of $s(x)$ copies of $g_x$, and let $C$ be a random sample of*

$$t = \frac{\dim(F) S^2}{\epsilon^2}$$

*functions from the set $G = \bigcup_{x \in \mathcal{X}} G_x$. Then for every $Q \in \mathcal{Q}$*

$$\left| \sum_{x \in \mathcal{X}} f_x(Q) - \sum_{c \in C} g_c(Q) \right| \leq \epsilon \sum_{x \in \mathcal{X}} f_x(Q).$$

# B ANALYSIS OF ALGORITHM 2

**Definition 6** *Let $\mathcal{X} \subset \mathbb{R}^d$ be a finite data set. A set $B \subset \mathbb{R}^d, |B| = \beta$ is an $(\alpha, \beta)$-bicriteria solution with respect to the optimal $k$-clustering with squared Mahalanobis distance $d_A$ iff*

$$\sum_{x \in \mathcal{X}} d_A(x, B) \leq \alpha \min_{\substack{C \subset \mathbb{R}^d \\ |C| = k}} \sum_{x \in \mathcal{X}} d_A(x, C).$$

**Theorem 5** *Let $k \in \mathbb{N}$. Let $\mathcal{X}$ be a finite set of points in $\mathbb{R}^d$ and $d_A$ be a squared Mahalanobis distance. Denote by $B$ the output of Algorithm 2. Then, with probability at least $1/2$, $B$ is a $(\alpha, \beta)$-bicriteria solution with*

$$\alpha = 16(\log_2 k + 2)$$

*and $\beta = k$. The probabilty can be boosted to $1 - \delta$ by running the algorithm $\log \frac{1}{\delta}$ times and selecting the solution with the lowest cost in terms of squared Mahalanobis distance. The time complexity of Algorithm 2 is $\mathcal{O}(nkd)$.*

**Proof** Since $d_A$ is a squared Mahalanobis distance, $A$ is symetric and positive definite. Hence, the Cholesky decomposition $A = U^T U$ is unique where $U$ is an upper triangular matrix. For all $p, q \in \mathbb{R}^d$

$$(p - q)^T A(p - q) = ||Up - Uq||_2^2.$$

As a result, the map $p \to Up$ in $\mathbb{R}^d \to \mathbb{R}^d$ is an isometry (distance preserving map) with regards to the metric spaces $(\mathbb{R}^d, \sqrt{d_A})$ and $(\mathbb{R}^d, \sqrt{d_{||\cdot||_2^2}})$. Furthermore, the isometry is bijective since $U$ is invertible.

As a direct consequence of the isometry, Algorithm 2 is equivalent to running $D^2$-sampling in the transformed space with the squared Euclidean distance $d_{||\cdot||_2^2}$. Consider the transformed data set $\tilde{\mathcal{X}} = \{Ux \mid x \in \mathcal{X}\}$ and the transformed solution $\tilde{B} = \{Ub \mid b \in B\}$. By Theorem 3.1 of Arthur and Vassilvitskii (2007), we have that

$$\mathbb{E} \left[ \sum_{\tilde{x} \in \tilde{\mathcal{X}}} d_{||\cdot||_2^2}(\tilde{x}, \tilde{B}) \right] \leq 8(\log_2 k + 2) \min_{\substack{\tilde{C} \subset \mathbb{R}^d \\ |\tilde{C}| = k}} \sum_{\tilde{x} \in \tilde{\mathcal{X}}} d_{||\cdot||_2^2}(\tilde{x}, \tilde{C}).$$

Markov's inequality implies that with probability at least $1/2$

$$\sum_{\tilde{x} \in \tilde{\mathcal{X}}} d_{||\cdot||_2^2}(\tilde{x}, \tilde{B}) \leq 16(\log_2 k + 2) \min_{\substack{\tilde{C} \subset \mathbb{R}^d \\ |\tilde{C}| = k}} \sum_{\tilde{x} \in \tilde{\mathcal{X}}} d_{||\cdot||_2^2}(\tilde{x}, \tilde{C}).$$

Due to the global isometry defined above, this implies

$$\sum_{x \in \mathcal{X}} d_A(x, B) \leq 16(\log_2 k + 2) \min_{\substack{C \subset \mathbb{R}^d \\ |C| = k}} d_A(x, C).$$

By construction, $|B| = k$ which concludes the proof. ∎

# C   HARD CLUSTERING

## C.1   Sensitivity

**Lemma 2** *Let* $d_\phi$ *be a* $\mu$-*similar Bregman divergence on domain* $\mathcal{K}$ *and denote by* $d_A$ *the corresponding squared Mahalanobis distance. Let* $\mathcal{X}$ *be a set of points in* $\mathcal{K}$ *and let* $B \subseteq \mathcal{K}$ *be an* $(\alpha, \beta)$-*bicriteria solution with respect to the optimal* $k$-*clustering with squared Mahalanobis distance* $d_A$. *For each point* $x \in \mathcal{X}$, *denote by* $b_x$ *the closest cluster center in* $B$ *in terms of* $d_A$ *and by* $\mathcal{X}_x$ *the set of all points* $x' \in \mathcal{X}$ *such that* $b_x = b_{x'}$. *Then, the sensitivity* $\sigma_{\mathcal{Q}}(x)$ *of the function*

$$f_x(Q) = \min_{q \in Q} d_\phi(x, q)$$

*is bounded for all* $x \in \mathcal{X}$ *by the function*

$$s(x) = \frac{4}{\mu} \left[ \frac{\alpha \, d_A(x, b_x)}{2 \bar{c}_B} + \frac{\alpha \sum_{x' \in \mathcal{X}_x} d_A(x', b_x)}{|\mathcal{X}_x| \bar{c}_B} + \frac{n}{|\mathcal{X}_x|} \right]$$

*where* $\bar{c}_B = \frac{1}{n} \sum_{x' \in \mathcal{X}} d_A(x', B)$. *Furthermore,*

$$S = \frac{1}{n} \sum_{x \in \mathcal{X}} s(x) = \frac{6\alpha + 4\beta}{\mu}.$$

## Proof

We consider an arbitrary point $x \in \mathcal{X}$ and an arbitrary query $Q \in \mathcal{Q}$ and define

$$\bar{c}_Q = \frac{1}{n} \sum_{x' \in \mathcal{X}} d_A(x', Q) \quad \text{and} \quad \bar{c}_B = \frac{1}{n} \sum_{x' \in \mathcal{X}} d_A(x', B).$$

Since $d_\phi$ is $\mu$-similar, we have

$$f_x(Q) = d_\phi(x, Q) \leq d_A(x, Q)$$

as well as $\bar{f}_Q \geq \mu \bar{c}_Q$. This implies

$$\frac{f_x(Q)}{\bar{f}_Q} \leq \frac{1}{\mu} \frac{d_A(x, Q)}{\bar{c}_Q}. \tag{4}$$

By the double triangle inequality, we have that

$$d_A(x, Q) \leq 2 \, d_A(x, b_x) + 2 \, d_A(b_x, Q)$$

which in combination with (4) implies

$$\frac{f_x(Q)}{\bar{f}_Q} \leq \frac{2}{\mu} \frac{d_A(x, b_x) + d_A(b_x, Q)}{\bar{c}_Q}. \tag{5}$$

Similarly, we have for all $x' \in \mathcal{X}_x$

$$d_A(b_x, Q) \leq 2 \, d_A(x', b_x) + 2 \, d_A(x', Q)$$

and thus

$$d_A(b_x, Q) \leq \frac{2}{|\mathcal{X}_x|} \sum_{x' \in \mathcal{X}_x} [d_A(x', b_x) + d_A(x', Q)].$$

Together with (5), this allows us to bound $\frac{f_x(Q)}{\bar{f}_Q}$ by

$$\frac{4}{\mu} \left[ \frac{d_A(x, b_x)}{2 \bar{c}_Q} + \frac{\sum_{x' \in \mathcal{X}_x} [d_A(x', b_x) + d_A(x', Q)]}{|\mathcal{X}_x| \bar{c}_Q} \right].$$

By definition, we have both $\bar{c}_Q \geq \frac{1}{\alpha} \bar{c}_B$ and $\bar{c}_Q \geq \frac{1}{|\mathcal{X}_x|} \sum_{x' \in \mathcal{X}_x} d_A(x', Q)$. This implies that

$$s(x) = \frac{4}{\mu} \left[ \frac{\alpha \, d_A(x, b_x)}{2 \bar{c}_B} + \frac{\alpha \sum_{x' \in \mathcal{X}_x} d_A(x', b_x)}{|\mathcal{X}_x| \bar{c}_B} + \frac{n}{|\mathcal{X}_x|} \right]$$

is a bound for the sensitivity $\sigma_{\mathcal{Q}}(x)$ since the choice of both $x \in \mathcal{X}$ and $Q \in \mathcal{Q}$ was arbitrary.

Using the definition of $\bar{c}_B$, we further have

$$S = \frac{1}{n} \sum_{x \in \mathcal{X}} s(x)$$

$$= \frac{1}{\mu} \left( 2\alpha + \frac{4}{n} \sum_{x \in \mathcal{X}} \left[ \frac{\alpha \sum_{x' \in \mathcal{X}_x} d_A(x', b_x)}{|\mathcal{X}_x| \bar{c}_B} + \frac{n}{|\mathcal{X}_x|} \right] \right)$$

$$= \frac{6\alpha + 4\beta}{\mu}.$$

which concludes the proof.   ∎

## C.2   Pseudo-dimension

**Theorem 6** *Let* $k \in \mathbb{N}$. *Let* $d_\phi$ *be a Bregman divergence on domain* $\mathcal{K} \subseteq \mathbb{R}^d$ *and* $\mathcal{X}$ *a finite set of points in* $\mathcal{K}$. *Define the set* $F = \{f_x(Q) \mid x \in \mathcal{X}\}$ *from* $\mathcal{K}^k$ *to* $[0, \infty)$ *where* $f_x(Q) = \min_{q \in Q} d_\phi(x, q)$. *Then, it holds that* $\dim(F) = (d + 2)k$.

**Proof** Consider an arbitrary subset $S \subseteq F$. We need to show

$$|\{S \cap R \mid R \in \mathbf{ranges}(F)\}| \leq |S|^d$$

which holds if

$$\left| \{ \text{range}(S, Q, r) \mid Q \in \mathbb{R}^{d \times k}, r \geq 0 \} \right| \leq |S|^d$$

where $\text{range}(S, Q, r) = \{f_x \in S \mid f_x(Q) \leq r\}$.

We first show the result for $k = 1$. For arbitrary $q \in \mathbb{R}^d$ and $r \geq 0$, we have

$$\{f_x \in S \mid f_x(\{q\}) \leq r\}$$
$$= \{f_x \in S \mid d_\phi(x, q) \leq r\}$$
$$= \{f_x \in S \mid \phi(x) - \phi(q) - \langle x - q, \nabla \phi(q) \rangle \leq r\}$$

As in Nielsen et al. (2007), we define the lifting map $\tilde{x} = [x, \phi(x)]$ and $\tilde{q} = [q, \phi(q)]$. Furthermore, we set $s = [-\nabla \phi(q), 1]$ and $t = r + \langle \tilde{q}, s \rangle$. We then have

$$\{f_x \in S \mid \phi(x) - \phi(q) - \langle x - q, \nabla \phi(q) \rangle \leq r\}$$
$$= \{f_x \in S \mid \langle \phi(x) - \phi(q), 1 \rangle + \langle x - q, -\nabla \phi(q) \rangle \leq r\}$$
$$= \{f_x \in S \mid \langle \tilde{x} - \tilde{q}, s \rangle \leq r\}$$
$$= \{f_x \in S \mid \langle \tilde{x}, s \rangle \leq t\}.$$

For every Bregman ball defined by $q$ and $r$, there is hence a corresponding halfspace in the lifted $d + 1$ dimensional space. We may obtain a bound on the pseudo-dimension of Bregman balls by bounding the pseudo-dimension of halfspaces. Using Theorem 3.1 of Anthony and Bartlett (2009), we have

$$\left| \left\{ \mathrm{range}(S, \{q\}, r) \mid q \in \mathbb{R}^d, r \geq 0 \right\} \right|$$
$$\leq 2 \sum_{l=0}^{d+1} \binom{|S| - 1}{l}$$
$$\leq \sum_{l=0}^{d+2} \binom{|S| - 1}{l} + \sum_{l=0}^{d+1} \binom{|S| - 1}{l}$$
$$= \sum_{l=0}^{d+2} \binom{|S|}{l} \leq \sum_{l=0}^{d+2} \binom{d+2}{l} |S|^l$$
$$= (|S| + 1)^{d+2}$$

which shows the claim for $k = 1$.

We now extend the result to $k \in \mathbb{N}$ centers. For arbitrary $Q \in \mathbb{R}^{d \times k}$ and $r \geq 0$, we have

$$\mathrm{range}(S, Q, r) = \{ f_x \in S \mid f_x(Q) \leq r \}$$
$$= \left\{ f_x \in S \mid \min_{q \in Q} \mathrm{d}_\phi(x, q) \leq r \right\}$$
$$= \bigcup_{q \in Q} \{ f_x \in S \mid \mathrm{d}_\phi(x, q) \leq r \}.$$

Hence,

$$\left| \left\{ \mathrm{range}(S, Q, r) \mid Q \in \mathbb{R}^{d \times k}, r \geq 0 \right\} \right|$$
$$\leq \left| \left\{ \mathrm{range}(S, \{q\}, r) \mid q \in \mathbb{R}^d, r \geq 0 \right\} \right|^k$$
$$\leq (|S| + 1)^{(d+2)k}$$

which concludes the proof since the choice of $S$ was arbitrary. ∎

### C.3 Proof of Theorem 1

**Theorem 1** *Let $\varepsilon \in (0, 1/4), \delta > 0$ and $k \in \mathbb{N}$. Let $\mathrm{d}_\phi$ be a $\mu$-similar Bregman divergence on domain $\mathcal{K}$ and denote by $\mathrm{d}_A$ the corresponding squared Mahalanobis distance. Let $\mathcal{X}$ be a set of points in $\mathcal{K}$ and let $B \subseteq \mathcal{X}$ be the set with the smallest quantization error in terms of $\mathrm{d}_A$ among $\mathcal{O}(\log \frac{1}{\delta})$ runs of Algorithm 2. Let $C$ be the output of Algorithm 3 with*

$$m = \mathcal{O}\left( \frac{dk^3 + k^2 \log \frac{1}{\delta}}{\mu^2 \varepsilon^2} \right).$$

*Then, with probability at least $1 - \delta$, the set $\mathcal{C}$ is a $(\varepsilon, k)$-coreset of $\mathcal{X}$ for hard clustering with $\mathrm{d}_\phi$.*

**Proof** Apply Theorem 5, Lemma 2 and Theorem 6 to Theorem 4. The results can be extended to hold with arbitrary probability $1 - \delta$ by Theorem 4.4 of Feldman and Langberg (2011). ∎

### C.4 Proof of Lemma 1

**Lemma 1** *Let $\epsilon \in (0, 1)$ and let $\mathrm{d}_\phi$ be a $\mu$-similar Bregman divergence on domain $\mathcal{K}$. Let $\mathcal{X} \subseteq \mathcal{K}$ be a data set, $k \in \mathbb{N}$ and $\mathcal{C}$ be an $(\varepsilon/3, k)$-coreset of $\mathcal{X}$ for hard clustering with $\mathrm{d}_\phi$. Let $Q^*_{\mathcal{X}}$ and $Q^*_{\mathcal{C}}$ denote the optimal set of cluster centers for $\mathcal{X}$ and $\mathcal{C}$ respectively. Then,*

$$\mathrm{cost}_{\mathrm{H}}(\mathcal{X}, Q^*_{\mathcal{C}}) \leq (1 + \epsilon) \, \mathrm{cost}_{\mathrm{H}}(\mathcal{X}, Q^*_{\mathcal{X}}).$$

**Proof** By the coreset property, we have

$$\mathrm{cost}_{\mathrm{H}}(\mathcal{X}, Q^*_{\mathcal{C}}) \leq \frac{1}{1 - \varepsilon/3} \, \mathrm{cost}_{\mathrm{H}}(\mathcal{C}, Q^*_{\mathcal{C}})$$
$$\leq \frac{1}{1 - \varepsilon/3} \, \mathrm{cost}_{\mathrm{H}}(\mathcal{C}, Q^*_{\mathcal{X}})$$
$$\leq \frac{1 + \varepsilon/3}{1 - \varepsilon/3} \, \mathrm{cost}_{\mathrm{H}}(\mathcal{X}, Q^*_{\mathcal{X}})$$
$$\leq (1 + \varepsilon) \, \mathrm{cost}_{\mathrm{H}}(\mathcal{X}, Q^*_{\mathcal{X}}).$$

∎

## D SOFT CLUSTERING

### D.1 Sensitivity

**Lemma 3** *Let $\mathrm{d}_\phi$ be a Bregman divergence that is $\mu$-similar on the set $\mathcal{K}$ and $\mathrm{d}_A$ denote the corresponding squared Mahalanobis distance. Denote by $\mathcal{X}$ a finite set of points in $\mathcal{K}$, $\Delta^k$ be the $k$-simplex and define $\mathcal{Q} = \Delta^k \times \mathcal{K}^k$. For $x \in \mathcal{X}$ and $Q = [w_1, \ldots, w_k, \theta_1, \ldots, \theta_k] \in \mathcal{Q}$ define*

$$f_\phi(x \mid Q) = -\ln\left( \sum_j w_j \exp\left(- \mathrm{d}_\phi(x, \theta_j)\right) \right)$$
$$f_A(x \mid Q) = -\ln\left( \sum_j w_j \exp\left(- \mathrm{d}_A(x, \theta_j)\right) \right)$$

*Then, for all $x \in \mathcal{X}$ and $Q \in \mathcal{Q}$:*

*i)* $f_\phi(x \mid Q) \geq 0.$

*ii)* $f_\phi(x \mid Q) \geq \mathrm{d}_\phi(x, Q).$

*iii)* $\mu f_A(x \mid Q) \leq f_\phi(x \mid Q) \leq f_A(x \mid Q).$

*iv)* $f_A(x \mid Q) \leq 2 \, \mathrm{d}_A(x, b) + 2 f_A(b \mid Q).$

**Proof**

i) For a fixed $x \in \mathcal{X}$, consider the discrete random variable $Y$ that takes value $\exp(-\,\mathrm{d}_\phi(x, \theta_j))$ with probability $w_j$, for $1 \leq j \leq k$, $\sum_{j=1}^k w_j = 1$. Clearly, $\mathbb{E}[Y] \leq 1$ which implies

$$f_\phi(x \mid Q) = -\ln(\mathbb{E}[Y]) \geq 0.$$

ii) Let $\mathrm{d}_\phi(x, Q) = \min_{\theta_j \in Q} \mathrm{d}_\phi(x, \theta_j)$. Since

$$\sum_{j=1}^k w_j \exp(-\,\mathrm{d}_\phi(x, \theta_j)) \leq \sum_{j=1}^k w_j \exp(-\,\mathrm{d}_\phi(x, Q))$$
$$\leq \exp(-\,\mathrm{d}_\phi(x, Q)) \sum_{j=1}^k w_j \leq \exp(-\,\mathrm{d}_\phi(x, Q)),$$

it follows that

$$f(x \mid Q) = -\ln\left(\sum_j w_j \exp(-\,\mathrm{d}_\phi(x, \theta_j))\right)$$
$$\geq -\ln(\exp(-\,\mathrm{d}_\phi(x, Q)))$$
$$= \mathrm{d}_\phi(x, Q).$$

iii) From $\mathrm{d}_\phi(x, \theta_j) \geq \mu\,\mathrm{d}_A(x, \theta_j)$ it follows that

$$\ln\left(\sum_{j=1}^k w_j \exp(-\,\mathrm{d}_\phi(x, \theta_j))\right)$$
$$\leq \ln\left(\sum_{j=1}^k w_j \exp(-\mu\,\mathrm{d}_A(x, \theta_j))\right)$$
$$= \ln\left(\sum_{j=1}^k w_j \left[\exp(-\,\mathrm{d}_A(x, \theta_j))\right]^\mu\right)$$
$$\leq \ln\left(\sum_{j=1}^k w_j \exp(-\,\mathrm{d}_A(x, \theta_j))\right)^\mu$$
$$= \mu \ln\left(\sum_{j=1}^k w_j \exp(-\,\mathrm{d}_A(x, \theta_j))\right)$$

by Jensen's inequality on $g(x) = x^\mu$ which is concave for $\mu \in (0, 1]$. Hence

$$f_\phi(x \mid Q) \geq -\mu \ln\left(\sum_{j=1}^k w_j \exp(-\,\mathrm{d}_A(x, \theta_j))\right)$$

which implies

$$\mu f_A(x \mid Q) \leq f_\phi(x \mid Q).$$

To prove the other direction note that

$$f_\phi(x \mid Q) = -\ln\left(\sum_{j=1}^k w_j \exp(-\,\mathrm{d}_\phi(x, \theta_j))\right)$$
$$\leq -\ln\left(\sum_{j=1}^k w_j \exp(-\,\mathrm{d}_A(x, \theta_j))\right)$$
$$= f_A(x \mid Q)$$

since $\mathrm{d}_\phi(x, \theta_j) \leq \mathrm{d}_A(x, \theta_j)$ by definition.

iv) By triangle inequality it holds that

$$-\ln\left(\sum_j w_j \exp(-\,\mathrm{d}_A(x, \theta_j))\right)$$
$$\leq -\ln\left(\sum_j w_j \exp(-2\,\mathrm{d}_A(x, b) - 2\,\mathrm{d}_A(b, \theta_j))\right)$$
$$= 2\,\mathrm{d}_A(x, b) - \ln\left(\sum_j w_j \left[\exp(-\,\mathrm{d}_A(b, \theta_j))\right]^2\right)$$
$$\leq 2\,\mathrm{d}_A(x, b) - 2\ln\left(\sum_j w_j \exp(-\,\mathrm{d}_A(b, \theta_j))\right)$$
$$= 2\,\mathrm{d}_A(x, b) + 2 f_A(b \mid Q),$$

by Jensen's inequality on $g(x) = x^2$. ∎

**Lemma 4** *Let $\mathrm{d}_\phi$ be a $\mu$-similar Bregman divergence on domain $\mathcal{K}$ and denote by $\mathrm{d}_A$ the corresponding squared Mahalanobis distance. Let $\mathcal{X}$ be a set of points in $\mathcal{K}$ and let $B \subseteq \mathcal{K}$ be an $(\alpha, \beta)$-bicriteria solution with respect to the optimal $k$-clustering with squared Mahalanobis distance $\mathrm{d}_A$. For each point $x \in \mathcal{X}$, denote by $b_x$ the closest cluster center in $B$ in terms of $\mathrm{d}_A$ and by $\mathcal{X}_x$ the set of all points $x'$ such that $b_x = b_{x'}$. Then, the sensitivity $\sigma_{\mathcal{Q}}(x)$ of the function*

$$f_x(Q) = f_\phi(x \mid Q) = -\ln\left(\sum_j w_j \exp(-\,\mathrm{d}_\phi(x, \mu_j))\right),$$

*is bounded for all $x \in \mathcal{X}$ by the function*

$$s(x) = \frac{4}{\mu}\left[\frac{\alpha\,\mathrm{d}_A(x, b_x)}{2\bar{c}_B} + \frac{\alpha \sum_{x' \in \mathcal{X}_x} \mathrm{d}_A(x', b_x)}{|\mathcal{X}_x|\bar{c}_B} + \frac{n}{|\mathcal{X}_x|}\right]$$

*where $\bar{c}_B = \frac{1}{n}\sum_{x' \in \mathcal{X}} \mathrm{d}_A(x', B)$. Furthermore,*

$$S = \frac{1}{n}\sum_{x \in \mathcal{X}} s(x) = \frac{1}{\mu}(6\alpha + 4\beta).$$

**Proof** Consider an arbitrary point $x \in \mathcal{X}$ and an arbitrary query $Q \in \mathcal{Q}$ and define

$$\bar{c}_Q = \frac{1}{n}\sum_{x' \in \mathcal{X}} \mathrm{d}_A(x', Q) \quad \text{and} \quad \bar{c}_B = \frac{1}{n}\sum_{x' \in \mathcal{X}} \mathrm{d}_A(x', B).$$

By Lemma 3 it holds that

$$\bar{f}_Q = \frac{1}{n} \sum_{x \in \mathcal{X}} f_x(Q) \geq \mu \frac{1}{n} \sum_{x \in \mathcal{X}} f_A(Q) = \mu \bar{c}_Q, \quad (6)$$

and $\forall x \in \mathcal{X}_x$

$$f_A(b_x \mid Q) \leq 2 \, \mathrm{d}_A(x, b_x) + 2 f_A(b_x \mid Q).$$

Summing over all $x \in \mathcal{X}_x$ yields

$$f_A(b_x \mid Q) \leq \frac{2}{|\mathcal{X}_x|} \sum_{x' \in \mathcal{X}_x} \left[ \mathrm{d}_A(x', b_x) + f_A(b_x \mid Q) \right]. \quad (7)$$

From Lemma 3 and (6) we can conclude that

$$\frac{f_x(Q)}{\bar{f}_Q} \leq \frac{1}{\mu} \frac{f_A(x \mid Q)}{\bar{c}_Q} \leq \frac{2}{\mu} \left[ \frac{\mathrm{d}_A(x, b_x)}{\bar{c}_Q} + \frac{f_A(b_x \mid Q)}{\bar{c}_Q} \right]$$

which combined with (7) suffices to bound $\frac{f_x(Q)}{\bar{f}_Q}$ by

$$\frac{4}{\mu} \left[ \frac{\mathrm{d}_A(x, b_x)}{2\bar{c}_Q} + \frac{\sum_{x' \in \mathcal{X}_x} \left[ \mathrm{d}_A(x', b_x) + f_A(b_x \mid Q) \right]}{\bar{c}_Q |\mathcal{X}_x|} \right].$$

By definition of $B$ and Lemma 3 it follows that

$$\bar{c}_Q = \frac{1}{n} \sum_{x' \in \mathcal{X}} f_A(x', Q) \geq \frac{1}{n} \sum_{x' \in \mathcal{X}} \mathrm{d}_A(x', Q) \geq \alpha \cdot \bar{c}_B$$

which implies that

$$s(x) = \frac{4}{\mu} \left[ \frac{\alpha \, \mathrm{d}_A(x, b_x)}{2\bar{c}_B} + \frac{\alpha \sum_{x' \in \mathcal{X}_x} \mathrm{d}_A(x', b_x)}{|\mathcal{X}_x| \bar{c}_B} + \frac{n}{|\mathcal{X}_x|} \right]$$

is a bound for the sensitivity $\sigma_{\mathcal{Q}}(x)$ since the choice of both $x \in \mathcal{X}$ and $Q \in \mathcal{Q}$ was arbitrary.

Finally, by substituting $\bar{c}_B$ in $s(x)$ it follows that

$$\begin{aligned}
S &= \frac{1}{n} \sum_{x \in \mathcal{X}} s(x) \\
&= \frac{1}{\mu} \left( 2\alpha + \frac{4}{n} \sum_{x \in \mathcal{X}} \left[ \frac{\alpha \sum_{x' \in \mathcal{X}} \mathrm{d}_A(x', b_x)}{|\mathcal{X}_x| \bar{c}_B} + \frac{n}{|\mathcal{X}_x|} \right] \right) \\
&= \frac{6\alpha + 4\beta}{\mu}.
\end{aligned}$$

which concludes the proof. ■

### D.2 Pseudo-dimension

To bound the pseudo-dimension of the function class we will first obtain a *solution set components bound*. Intuitively, we partition the parameter space into connected components where parameters from the same connected component realize the same dichotomy.

Hence, we can upper bound the number of possible dichotomies by upper bounding the number of connected components in the parameter space. More information on this method is available at Anthony and Bartlett (2009) (Chapter 7). We state the necessary Lemmas from Anthony and Bartlett (2009) and Schmitt (2002).

**Lemma 5** *Let $k$ be a natural number and suppose $G$ is the class of real-valued functions in the variables $y_1, \ldots, y_d$ satisfying the following conditions: For every $f \in G$ there exist affine functions $g_1, \ldots, g_k$ in the variables $y_1, \ldots, y_d$ such that $f$ is an affine combination of $y_1, \ldots, y_d$ and $e^{g_1}, \ldots, e^{g_k}$. Then $G$ has solution set components bound*

$$\mathcal{O}\left( 2^{d^2 k^2} \right).$$

**Definition 7** *A class of functions $\mathcal{F}$ is closed under addition of constants if for every $c \in \mathbb{R}$ and $f \in \mathcal{F}$ the function $z \to f(z) + c$ is a member of $\mathcal{F}$.*

The following Lemma is due to Schmitt (2002) and slightly improves on the result from Anthony and Bartlett (2009).

**Lemma 6** *Let $\mathcal{F}$ be a class of real-valued functions*

$$(y_1, \ldots, y_d, x_1, \ldots, x_n) \to f(y_1, \ldots, y_d, x_1, \ldots, x_n)$$

*that is closed under addition of constants and where each function in $\mathcal{F}$ is $C^d$ in the variables $y_1, \ldots, y_d$. If the class*

$$\mathcal{G} = \{(y_1, \ldots, y_d) \to f(y_1, \ldots, y_d, s) : f \in \mathcal{F}, s \in \mathbb{R}^n\}$$

*has solution set components bound $B$ then for any sets $\{f_1, \ldots, f_k\} \subseteq \mathcal{F}$ and $\{s_1, \ldots, s_m\} \subseteq \mathbb{R}^n$, where $m \geq d/k$, the set $T \subseteq \{0, 1\}^{mk}$ defined as*

$$\begin{aligned}
T = \{ & (\mathrm{sgn}(f_1(a, s_1)), \ldots, \mathrm{sgn}(f_1(a, s_m)), \\
& \mathrm{sgn}(f_2(a, s_1)), \ldots, \mathrm{sgn}(f_2(a, s_m)), \ldots, \\
& \mathrm{sgn}(f_k(a, s_1)), \ldots, \mathrm{sgn}(f_k(a, s_m)) : a \in \mathbb{R}^d \}
\end{aligned}$$

*satisfies*

$$|T| \leq B \sum_{i=0}^{d} \binom{mk}{i} \leq B \left( \frac{emk}{d} \right)^d.$$

We now prove the following Lemma which is necessary to bound the pseudo-dimension.

**Lemma 7** *Let $\mathbf{w} = (w_0, \ldots, w_k) \subset \mathbb{R}^{k+1}$, $y = (y_{11}, \ldots, y_{1d}, \ldots, y_{kd}) \subset \mathbb{R}^{kd}$ and $x = (x_1, \ldots, x_d) \subset \mathbb{R}^d$. Define*

$$f(\mathbf{w}, \mathbf{y}, \mathbf{x}) = w_0 + \sum_{j=1}^{k} w_j \exp\left( \sum_{i=1}^{d} y_{ji} x_i \right),$$

*and let $\mathcal{F} = \{f(\mathbf{w}, \mathbf{y}, \cdot) \mid \mathbf{w} \in \mathbb{R}^{k+1}, \mathbf{y} \subset \mathbb{R}^{kd}\}$. Then $\dim(\mathcal{F}) = \mathcal{O}(k^4 d^2)$.*

**Proof** Following Schmitt (2002) and Anthony and Bartlett (2009) we partition the functions $f \in \mathcal{F}$ into categories based on whether $w_i = 0$ or $w_i > 0$, $1 \leq i \leq k$, which results in $2^k$ categories. For each category we introduce new variables $w_1^\star, \ldots, w_k^\star$ where

$$w_i^\star = \begin{cases} \ln w_i & \text{if } w_i > 0 \\ 0 & \text{otherwise} \end{cases}$$

for $1 \leq i \leq k$. Choose an arbitrary category. Within this category the functions $f$ for an input $x \in \mathbb{R}^d$ can be expressed in the form

$$(w_0, w^\star, \theta) \to w_0 + b_1 e^{w_1^\star + \theta_1^T x} + \cdots + b_k e^{w_k^\star + \theta_k^T x}$$

where

$$b_i = \begin{cases} 0 & \text{if } w_i^\star = 0, \\ 1 & \text{otherwise.} \end{cases} \quad 1 \leq i \leq k.$$

Let $\mathcal{I} = \{(x_1, u_1), \ldots, (x_m, u_m)\}$ be an arbitrary set where $x_1, \ldots, x_m$ are input vectors in $\mathbb{R}^d$ and $u_1, \ldots, u_m$ are real numbers. We will estimate the number of dichotomies that are induced on set $\mathcal{I}$ by functions of the form

$$(x, z) \to \text{sgn}(f(x) - z).$$

Let $T \subseteq \mathbb{R}^{mk}$ contain all dichotomies induced by functions $f$ from a fixed category on a set $\mathcal{I}$

$$T = \{(\text{sgn}(f_1(a, x_1, u_1)), \ldots, \\ \text{sgn}(f_k(a, x_m, u_m)) : a \in \mathbb{R}^W\}$$

where each $f_i$ is of the form

$$(y, x, z) \to c_0 + y_0 + c_1 e^{y_1 + y_{11} x_{11} + \cdots + y_{1d} x_{1d}} + \ldots \quad (8) \\ + c_k e^{y_k + y_{k1} x_{k1} + \cdots + y_{kd} x_{kd}} - z.$$

The sets of variables $y_i$ and $y_{ij}$ play the role of the function parameters, $x_{ij}$ are the inputs and $z$ is the input variable for $u_1, \ldots, u_m$. Let $\mathcal{F}$ denote the class of the functions arising for real numbers $c_0, c_1, \ldots, c_k$ with $c_0 = 0$ and $c_i \in \{0, 1\}$ for $i = 1, \ldots, k$. We have introduced $c_0$ to make $\mathcal{F}$ closed under addition of constants. Now, for the vectors $x_1, \ldots, x_m$ and real numbers $u_1, \ldots, u_m$ consider the function class

$$\mathcal{G} = \{y \to f(y, x_i, u_i) : f \in \mathcal{F}, i = 1, \ldots, m\}.$$

Upon inspection of equation (8) we can see that every $f \in \mathcal{G}$ has $k$ exponents that are affine functions in $d$ variables. By Lemma 5 the class $\mathcal{G}$ has solution set components bound

$$B = \mathcal{O}\left(2^{W^2 k^2}\right).$$

Since $\mathcal{F}$ is closed under addition of constants, we have from Lemma 6 that

$$|T| \leq B(emk/W)^W$$

which is by construction an upper bound on the number of dichotomies that are induced on any set of $m$ vectors $\{(x_1, u_1), \ldots, (x_m, u_m)\}$. Since the choice of the category was arbitrary, it follows that

$$|T| \leq B(emk/W)^W 2^k,$$

by considering all $2^k$ categories. By definition, shattering $m$ vectors implies that all $2^m$ dichotomies must be induced. Since $T$ contains all the induced dichotomies it must hold that

$$m \leq \log B + W \log(emk/W) + k \log 2.$$

Using the fact that $\ln \alpha \leq \alpha\beta + \ln(1/\beta) - 1, \alpha, \beta > 0$ with $\alpha = m$ and $\beta = \frac{\ln 2}{2W}$ we obtain

$$W \log m \leq \frac{m}{2} + W \log \frac{2W}{e \ln 2},$$

which implies

$$m \leq 2 \log B + 2W \log \frac{2k}{\ln 2} + 2k \log 2.$$

Substituting the solution set components bound $B$ we conclude that

$$m = \mathcal{O}\left(W^2 k^2 + Wk \log(Wk) + W \log k\right) = \mathcal{O}\left(W^2 k^2\right).$$

The number of parameters $W$ is $\mathcal{O}(kd + k) = \mathcal{O}(kd)$ which implies $\dim(\mathcal{F}) = \mathcal{O}\left(k^4 d^2\right)$ as claimed. ■

Finally, we are ready to prove the main result by applying Lemma 7 on a suitable reparametrization of the Exponential family.

**Theorem 7** *Let $k \in \mathbb{N}$. Let $d_\phi$ be a Bregman divergence on domain $\mathcal{K} \subseteq \mathbb{R}^d$ and $\mathcal{X}$ a finite set of points in $\mathcal{K}$. Let $\mathcal{F} = \{f_x(Q) \mid x \in \mathcal{X}\}$ where*

$$f_x(Q) = f_\phi(x \mid Q) = -\ln\left(\sum_{j=1}^{k} w_j \exp\left(-d_\phi(x, \mu_j)\right)\right).$$

*Then, it holds that $\dim(\mathcal{F}) = \mathcal{O}\left(k^4 d^2\right)$.*

**Proof** We have that $\forall x \in dom(\phi)$

$$\exp(-d_\phi(x, \mu)) b_\psi(x) = \exp(x^T \theta - \psi(\theta) - g_\psi(x)) \\ = \exp(\bar{x}^T \bar{\theta})$$

where $x$ is the sufficient statistic, $\theta$ is the natural parameter, $\mu$ is the corresponding expectation parameter, $\psi(\theta)$ is the cumulant function, $\bar{x} = [x, -1, -1]$, and $\bar{\theta} = [\theta, -\psi(\theta), -g_\psi(x)]$. Hence,

$$f_\phi(x \mid Q) = -\ln\left(\sum_{j=1}^{k} w_j \exp\left(-\mathrm{d}_\phi(x, \mu_j)\right)\right)$$
$$= -\ln\left(\sum_{j=1}^{k} w_j \exp\left(\bar{x}^T \bar{\theta}_j\right)\right).$$

By Lemma 7 the function inside the natural logarithm has pseudo-dimension of $\mathcal{O}(k^4 d^2)$. The upper bound on the pseudo-dimension is preserved under monotonic transformations (Anthony and Bartlett, 2009) which concludes the proof. ∎

### D.3 Proof of Theorem 3

**Theorem 3** *Let $\varepsilon \in (0, 1/4), \delta > 0$ and $k \in \mathbb{N}$. Let $\mathrm{d}_\phi$ be a $\mu$-similar Bregman divergence on domain $\mathcal{K}$ and denote by $\mathrm{d}_A$ the corresponding squared Mahalanobis distance. Denote by $\mathcal{X}$ a set of points in $\mathcal{K}$. Let $\mathcal{C}$ be the output of Algorithm 3 with coreset size*

$$m = \mathcal{O}\left(\frac{d^2 k^4 + k^2 \log \frac{1}{\delta}}{\mu^2 \varepsilon^2}\right).$$

*Then, with probability at least $1-\delta$ the set $\mathcal{C}$ is a $(\varepsilon, k)$-coreset of $\mathcal{X}$ for soft clustering with $\mathrm{d}_\phi$.*

**Proof** Apply Theorem 5, Lemma 4 and Lemma 7 to Theorem 4. The results can be extended to hold with arbitrary probability $1 - \delta$ by Theorem 4.4 of Feldman and Langberg (2011). ∎